\documentclass[conference]{IEEEtran}
\IEEEoverridecommandlockouts
\usepackage{cite}
\usepackage{amsmath,amssymb,amsfonts}
\usepackage{graphicx}
\usepackage{textcomp}
\usepackage{xcolor}
\usepackage{graphicx}
\usepackage{booktabs}
\usepackage{makecell}
\usepackage{multirow}
\usepackage{xcolor}
\usepackage{algpseudocode}

\def\BibTeX{{\rm B\kern-.05em{\sc i\kern-.025em b}\kern-.08em
    T\kern-.1667em\lower.7ex\hbox{E}\kern-.125emX}}
\begin{document}

\title{Discovering the Effectiveness of Pre-Training \\in a Large-scale Car-sharing Platform
}

\author{Kyung Ho Park$^{1}$ and Hyunhee Chung$^{1}$
\thanks{$^{1}$K. H. Park is with SOCAR AI Research.
        {\tt\small kp at socar.kr}}%
\thanks{$^{1}$H. Chung is with SOCAR AI Research.
        {\tt\small esther at socar.kr}}%
}

\maketitle

\begin{abstract}
Recent progress of deep learning has empowered various intelligent transportation applications, especially in car-sharing platforms. While the traditional operations of the car-sharing service highly relied on human engagements in fleet management, modern car-sharing platforms let users upload car images before and after their use to inspect the cars without a physical visit. To automate the aforementioned inspection task, prior approaches utilized deep neural networks. They commonly employed pre-training, a de-facto technique to establish an effective model under the limited number of labeled datasets. As candidate practitioners who deal with car images would presumably get suffered from the lack of a labeled dataset, we analyzed a sophisticated analogy into the effectiveness of pre-training is important. However, prior studies primarily shed a little spotlight on the effectiveness of pre-training. Motivated by the aforementioned lack of analysis, our study proposes a series of analyses to unveil the effectiveness of various pre-training methods in image recognition tasks at the car-sharing platform. We set two real-world image recognition tasks in the car-sharing platform in a live service, established them under the many-shot and few-shot problem settings, and scrutinized which pre-training method accomplishes the most effective performance in which setting. Furthermore, we analyzed how does the pre-training and fine-tuning convey different knowledge to the neural networks for a precise understanding. 

\end{abstract}


\section{Introduction}
\begin{figure}[ht]
\centering
\begin{tabular} {c} \\
\includegraphics[width=0.4\textwidth, trim={0cm 0cm 0cm 0cm}, clip]{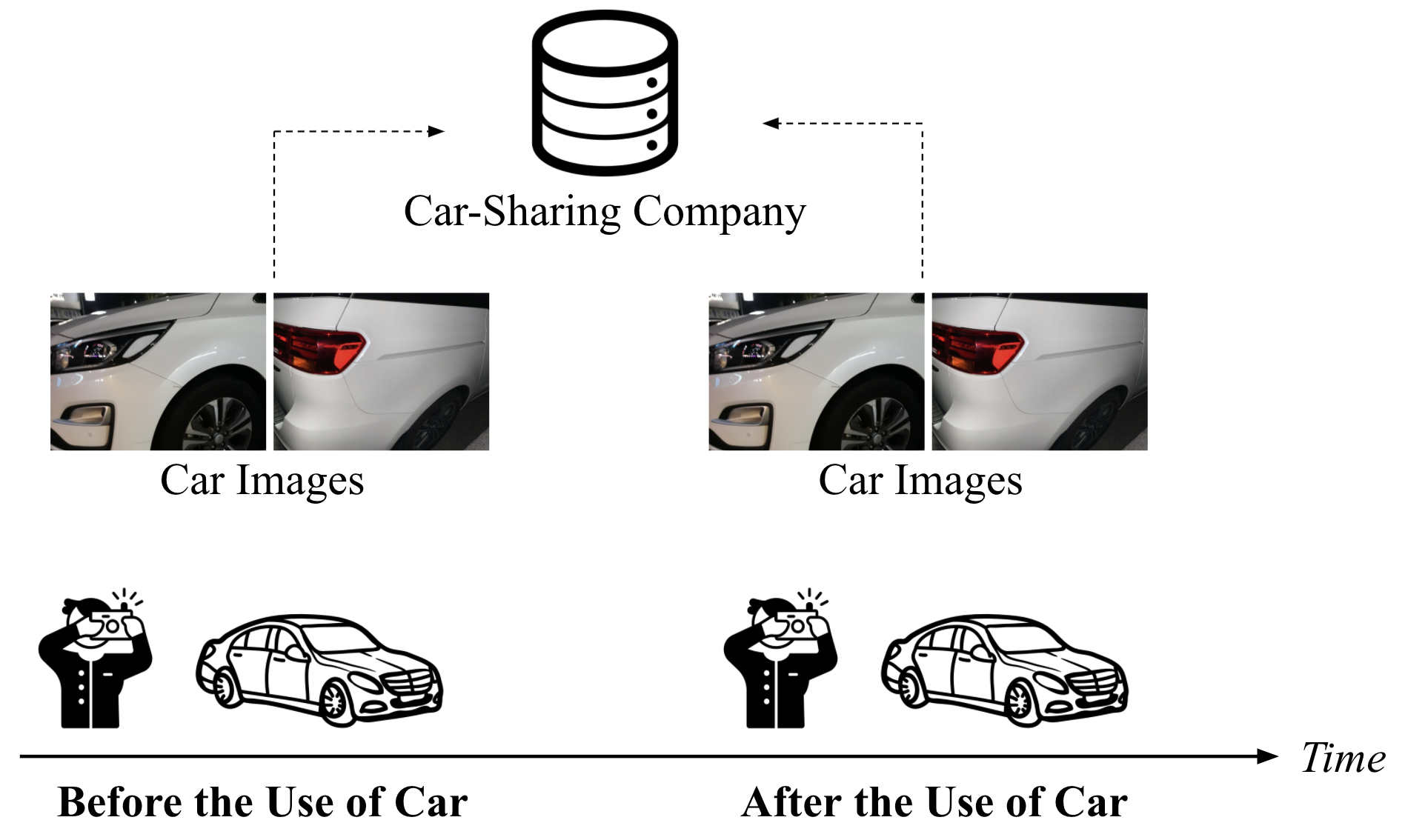} \\

\end{tabular}
\caption{Illustration of how car images are retrieved from the customers.}
\label{fig:example}
\end{figure}

The recent development of deep learning has empowered convenience and efficiency toward various areas of the transportation industry, such as traffic surveillance \cite{datondji2016survey,fang2016fine}, car insurance, and especially car-sharing services \cite{patil2017deep,singh2019automating}. In the past decades, the traditional car-sharing company's operational procedures mostly necessitated human engagement. Human operators should have resided in the rental station to make a rental contract with the customers, and they manually checked every car's condition for quality assurance; thus, the car-sharing business was known to necessitate a large amount of human labor. On the other hand, modern car-sharing services such as Zipcar in the United States have reduced the aforementioned burdensome human engagements in their operations, leveraging mobile technologies. As most people have smartphones, car-sharing companies started to let customers conveniently reserve the car and open the car's door through the smartphone application \cite{park2021visual,park2022don}. Moreover, as shown in Figure. \ref{fig:example}, the company requires the customers to take pictures of the car before and after its use to ensure that they did not damage it. The human operators inspect retrieved images to monitor the cars' condition without visiting the station. Throughout the aforementioned efforts, modern car-sharing companies aim to provide a convenient experience to the customers and accomplish efficiency in fleet management.

To automate the image-based fleet management procedures, several studies started to automate fleet management procedures with computer vision algorithms. Along with the development of deep neural networks, recent studies primarily employ convolutional neural networks (CNN) under the fully supervised, high data availability regime \cite{patil2017deep}. However, the prior supervised learning-based approaches bear several limits. First, the practitioner should manually annotate every image, and it creates a burdensome resource consumption. Second, especially in the car-sharing service, there frequently exists the case where the practitioner cannot acquire a particular amount of samples. 

Among several mitigating methods against these limits, pre-training has become a de-facto technique to establish a practical model in the real world. Given a target task where the amount of training samples is limited or the practitioner cannot annotate many samples, the pre-training method trains the feature extractor of the neural networks on the other task (source task) and fine-tunes the model on the target task \cite{pan2009survey}. One paradigm of pre-training is transfer learning, which trains the network on the external source dataset such as ImageNet \cite{deng2009imagenet} under the supervised manner. The other paradigm is self-supervised learning, which trains the network on the source task consisting of the target dataset in an unsupervised manner \cite{jing2020self}. These pre-training approaches are known to convey powerful representation power to the neural network \cite{pan2009survey}; thus, it enables the model to hedge the risk of overfitting under the limited number of samples at the target training set. While the image recognition studies related to the operations of car-sharing services conventionally utilized pre-training, we analyzed there has been less spotlight on the following questions: \textit{1) Which pre-training method is most effective at various problem settings to learn the representation power of car images?},  \textit{2) If an effective pre-training method exists, why does it contribute to better performance?}

To provide meaningful takeaways for the candidate practitioners who deal with car images, our study proposes a series of analyses to unveil the effectiveness of two pre-training paradigms in various image recognition tasks at the car-sharing service. Under the cooperation with the SOCAR, which is the largest car-sharing platform in Korea, we present two real-world image recognition problems: \textit{car model recognition} and \textit{car defect recognition}. To examine the most effective pre-training method in various problem settings, we established two learning paradigms: many-shot learning and few-shot learning. 
Throughout solving the aforementioned image recognition tasks in various problem settings, we discovered meaningful takeaways related to the effectiveness of pre-training. Furthermore, we scrutinized how pre-training methods convey knowledge to the target task, and which layers are affected by the pre-training and the fine-tuning. We highly expect the candidate practitioners can utilize these takeaways to efficiently solve real-world problems related to the image recognition tasks with car images.

    
    
    

\section{Related Works}

\subsection{Understanding Car-related Images}
Recently-proposed car-related studies aim to understand various attributes of the car with deep neural networks \cite{azimjonov2021real,azimjonov2022vision}. As early works, car part recognition methods have been proposed. Prior works resolved this task under the supervised regime, which trains the model with a finely-labeled large dataset. \cite{patil2017deep} acquired car images, annotated corresponding car parts, and trained CNNs under various architectures. \cite{singh2019automating} additionally merge both classifier and object detector to precisely understand car images. \cite{park2021visual} also employed an unsupervised learning paradigm to cluster car images including similar parts. Not only understand car parts, but several studies also proposed various studies understanding various car statuses. They suggested numerous approaches to understanding damaged patterns of the car to automate post-accident claims of the car-insurance entities \cite{park2022don}. Throughout these works, we confirm that cars have become an important object in modern computer vision studies, and both academia and industry have a keen interest in discovering effective representations of car-related images. 
However, we scrutinize most prior works of application studies to solve particular tasks (i.e., defect recognition), not focusing on why a particular learning strategy effectively understands car images. 
Furthermore, most previous studies utilize web-crawled or surveillance camera-taken images, which do not reflect the realistic characteristics of the car in the real world. Under the aforementioned limits of prior works, our study aims to focus on effective learning strategies regarding real-world car images and how a particular strategy contributes to the improved inductive biases of the model.

\subsection{Representation Learning}

Transfer learning aims to utilize the knowledge gained from solving the source task to escalate the target task performance \cite{pan2009survey}. 
A conventional method of transfer learning utilizes the feature extractor weight generated from the source task of the ILSVRC task (i.e., ImageNet classification task) \cite{deng2009imagenet}. We employed two transfer learning methods: a pre-training with the ImageNet dataset and a pre-training with Stanford-Cars dataset \cite{KrauseStarkDengFei-Fei_3DRR2013}. We employed ImageNet pre-trained weight as it has been conventionally utilized in numerous computer vision tasks due to its generality. The Stanford-Cars dataset is a dataset of cars; thus, we expected utilization of the Stanford-Cars dataset as a source dataset would share similar knowledge to our image recognition tasks.

Self-supervised learning utilizes Training set  during the pre-training and trains any human-annotated tasks \cite{jing2020self,jaiswal2021survey}. 
There have been proposed a wide range of pretext tasks for the source task, such as jigsaw puzzles \cite{noroozi2016unsupervised}, image colorization \cite{caron2020unsupervised}, or rotation prediction \cite{gidaris2018unsupervised}. One recent paradigm of self-supervised learning is contrastive learning \cite{jaiswal2021survey}, which trains the model to discriminate whether a pair of the original image and the augmented one come from the same source (i.e., SimCLR \cite{chen2020simple}, BYOL \cite{grill2020bootstrap}). 
While contrastive learning approaches accomplished precise performance, it bears a limit in that the model necessitates a massive amount of GPU computing environments \cite{chen2020simple}. As our study aims to provide meaningful takeaways that general practitioners can utilize, we dropped instance discrimination-based self-supervised learning approaches. Instead, our study employed a rotation prediction-based self-supervised learning approach \cite{gidaris2018unsupervised}, which achieved a promising performance in the public benchmark dataset without a huge computing environment.

\section{Preliminaries}
\subsection{Pre-Training Methods} \label{options}

Throughout the study, we employed four pre-training methods in various problem settings as below:

\begin{itemize}
    \item \textbf{Random}: It randomly initializes target feature extractor $F_T$'s weight. It implies the target neural network $N_T$ solves the target task from the randomly initialized feature extractors and classification layers without any pre-training.
    
    \item \textbf{TL-ImageNet}: Under the regime of transfer learning, it initializes target feature extractor $F_T$'s weight from the trained feature extractor $F_S$ trained on the ImageNet dataset. 
    
    \item \textbf{TL-Cars}: Under the regime of transfer learning, it initializes target feature extractor $F_T$'s weight from the trained feature extractor $F_S$ trained on the Stanford-Cars dataset. As the Stanford-Cars dataset and our image recognition tasks simultaneously deal with car images, we expect this option to provide contextual knowledge to the target neural network $N_T$.
    
    \item \textbf{SSL-Rotation}: Under the regime of self-supervised learning, it initializes target feature extractor $F_T$'s weight from the trained feature extractor $F_S$ trained from the rotation prediction as a pretext task. 
\end{itemize}

    
    
    

\begin{figure*}
\centering
\begin{tabular} {c c} \\
\includegraphics[width=0.45\textwidth]{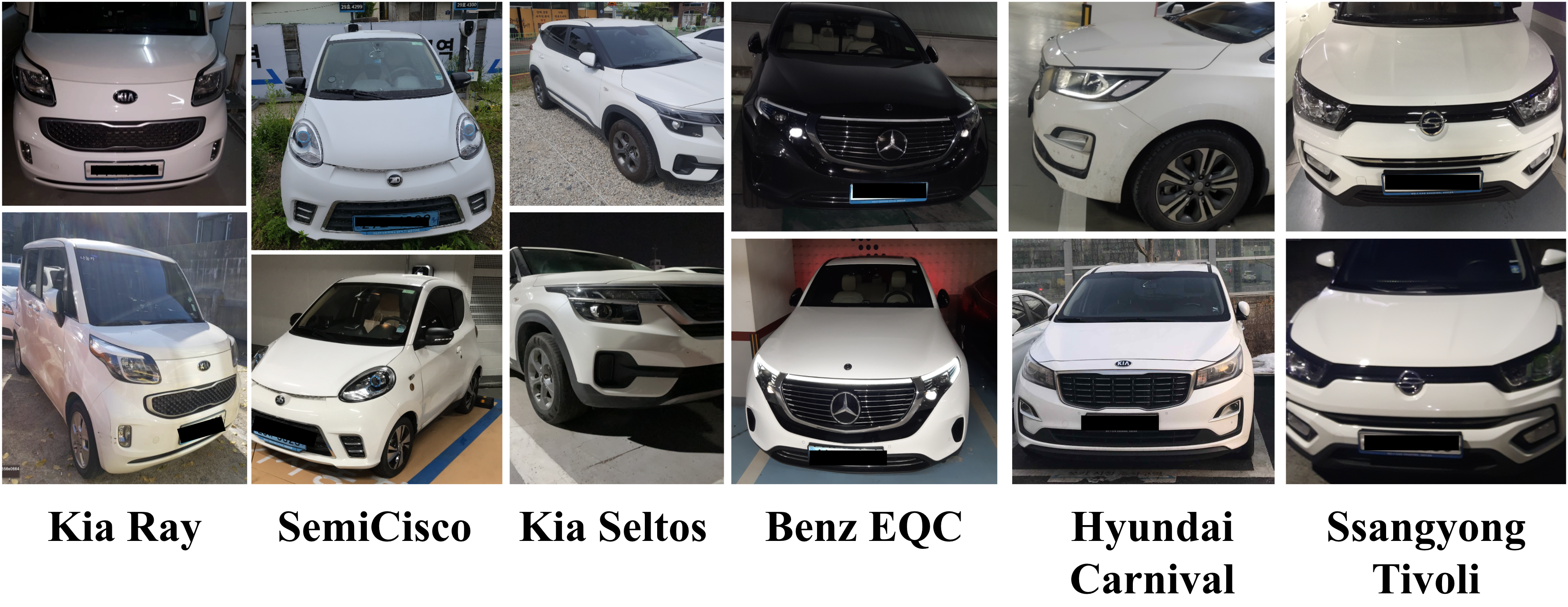} &
\includegraphics[width=0.42\textwidth]{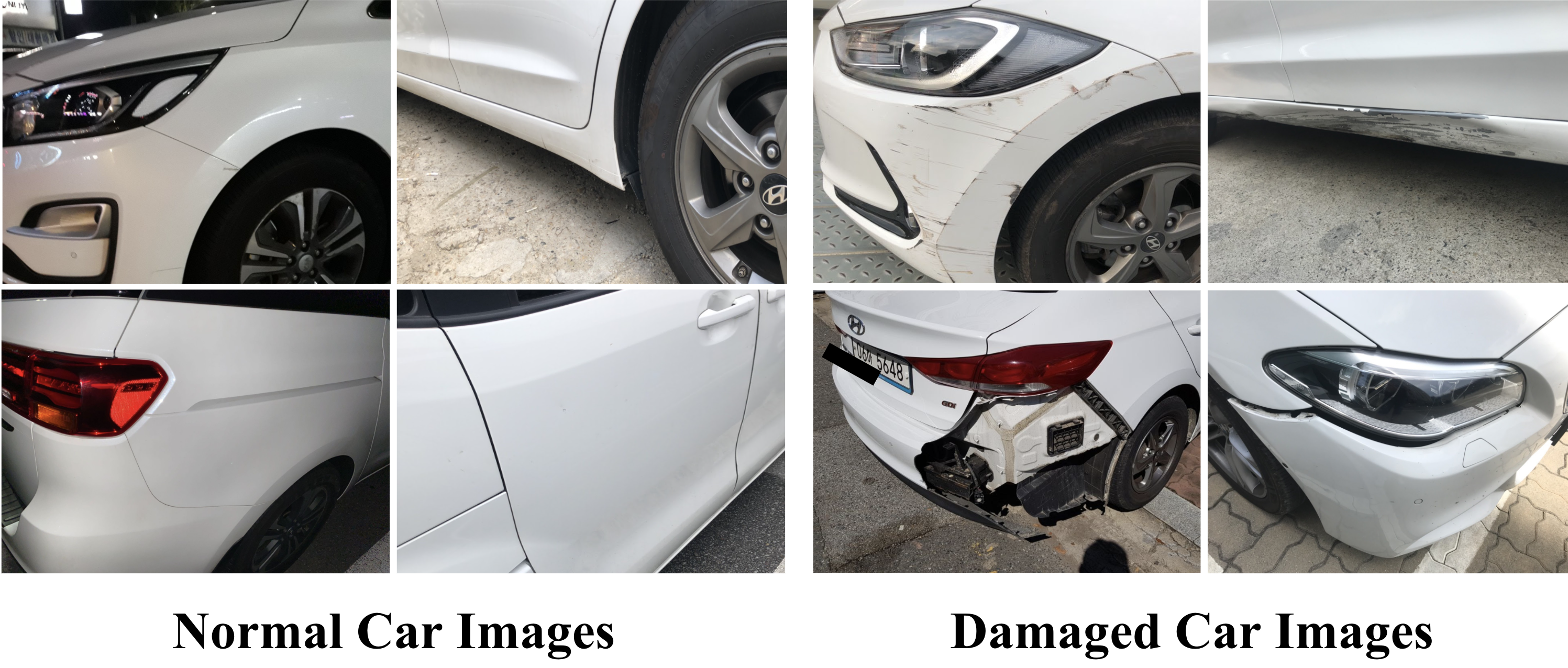} \\
(a) Car Model Dataset & (b) Car Defect Dataset \\
\end{tabular}
\caption{Sample images at \textit{car model dataset} and \textit{car defect dataset}}
\label{fig:sample_images}
\end{figure*}

\subsection{Target Tasks}
Our study employed two image recognition tasks that are widely utilized in real-world transportation industries: \textit{car model recognition} and \textit{car defect recognition}. The datasets for both tasks are accumulated from May 2021 to August 2021 from SOCAR, which is the largest car-sharing platform in Korea. As every image in both datasets is taken by real-world customers, sizes and resolutions vary at each sample. 

\subsubsection{Car model recognition}
Car model recognition is a $N$-class classification task given $N$ car models. The car model recognition model is an effective tool for internal human operators or data analysts at the car-sharing platform. When operators have to categorize a large number of unlabeled car images, they can utilize the classified results from the car model recognition model instead of manual inspections in every sample. The car model recognition model is also utilized to identify whether a customer correctly uploads the image of the car he or she borrowed, not the other car. 
In our study, we acquired \textit{car model dataset} consisting of 10 car models (2 light cars, 4 mid-sized cars, 4 Sport Utility Vehicles) for the car model recognition task, and samples from particular car models are visualized in Figure \ref{fig:sample_images}-(a). 

\subsubsection{Car Defect Recognition}
Car defect recognition is a binary classification task between normal car images and damaged car images. The car defect recognition model is primarily utilized for fleet management to ensure whether a car has a damaged part or not. If the model figures out a damaged car, human operators in the car-sharing platform visit the rental station to repair it. Our study retrieved \textit{car defect dataset}, which consists of both normal car images and damaged car images as shown in Figure  \ref{fig:sample_images}-(b). While the normal car images have a common characteristic of a clean surface, the damaged car images bear various damage types such as scratch, dent, separation of car parts, and severe crashes.

\begin{figure*}
\centering
\begin{tabular} {c c c} \\
\includegraphics[width=0.25\textwidth]{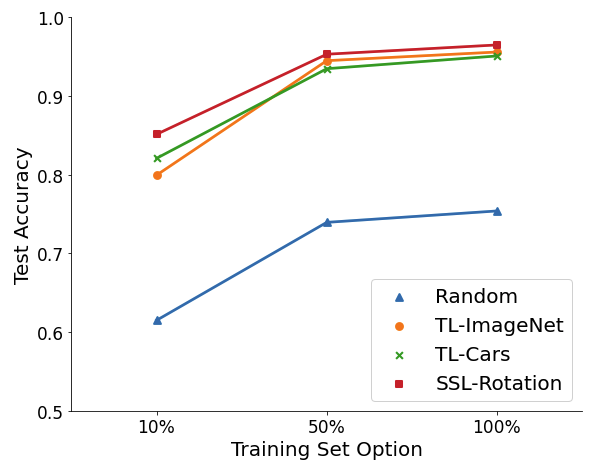} & 
\includegraphics[width=0.25\textwidth]{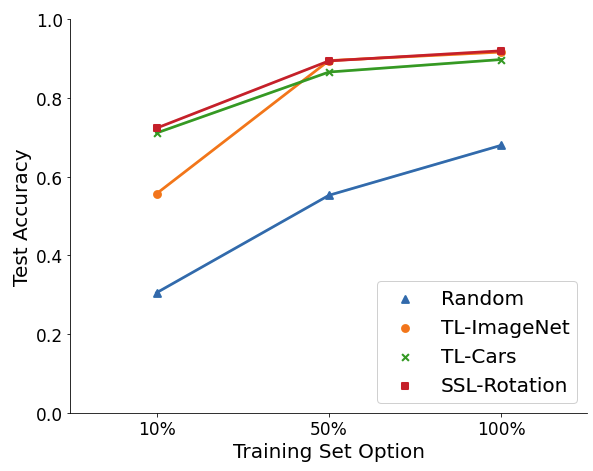} & 
\includegraphics[width=0.25\textwidth]{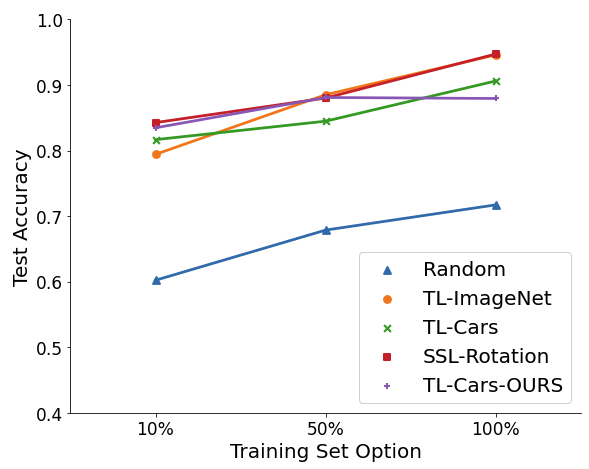} \\ 
(a) 2-Class Car Model Recognition & (b) 10-class Car Model Recognition & (c) Car Defect Recognition \\
\end{tabular}
\caption{Results on the effectiveness of various pre-training methods at \textit{car model recogntion} and \textit{car defect recognition} at the many-shot problem setting. 
}
\label{result:many}
\end{figure*}

\section{Effectiveness of Pre-Training under Many-shot setting} \label{sec:many}

\subsection{Problem Definition}
First and foremost, we examined how the pre-training methods escalate the target performance under the many-shot learning paradigm. Many-shot learning is a supervised learning paradigm where a particular amount of samples are available in the training set. To examine the effectiveness of various pre-training methods at two target tasks above, we established several research questions to provide essential takeaways toward candidate practitioners. The research questions are illustrated as follows.

\begin{itemize}
    \item \textbf{RQ1}: Does the pre-training contribute to the escalation of target task performance?

    \item \textbf{RQ2}: Does the effectiveness of pre-training differs at different sizes of the target training set? If it differs, which pre-training method is optimal in which size of training set?
\end{itemize}

\subsection{Many-shot Car Model Recognition} \label{many:model}
\subsubsection{Setup}
To validate the effectiveness of pre-training in car model recognition, we established two classification tasks with a car model dataset consisting of 10 car models (classes): 2-class classification and 10-class classification. We established two classification tasks to validate whether the pre-training method escalates the target performance well regardless of the number of classes. In response to the \textbf{RQ1}, we initialized the target feature extractor $F_T$ with four pre-training options proposed in the section \ref{options} and trained the $N_T$ with the target dataset $D_T$. We aim to figure out three pre-training options (\textit{TL-ImageNet}, \textit{TL-Cars}, \textit{SSL-Rotation}) to elevate the target task performance compared to the random initialization (Random). Moreover, we configured three types of the target training set with a different number of samples at each class to make a response to the \textbf{RQ2}: \textit{Model-10\%}, \textit{Model-50\%}, \textit{Model-100\%}. The \textit{Model-10\%} training set option implies a circumstance where the practitioners could not acquire many annotated samples at each class, and we expected this circumstance to frequently happen in the real world. The \textit{Model-100\%} training set illustrates the case where the practitioners acquired a comparatively large amount of labeled samples. Note that detailed training set options are illustrated in Table \ref{car_model_set}.

As an implementation detail, we employed the deep neural networks architecture of ResNet-50 \cite{he2016deep} both in pre-training and the target task as it is a widely-utilized network architecture in various computer vision tasks \cite{khan2020survey,he2016deep}. We employed a cross-entropy loss as a learning objective, Adam \cite{kingma2014adam} as an optimizer, and applied weight decay \cite{krogh1992simple} to accomplish the convergence during the target training stage. Note that we established pre-trained feature extractor weights $F_S$ on the ResNet-50 architecture before training the target model $N_T$, and we also checked the convergence during the pre-training. Lastly, we employed an evaluation metric of the target task performance as classification accuracy. Based on the aforementioned setups, we checked the target task performance along with different pre-training methods and target training set options. The experiment results at 2-class classification and 10-class classification are illustrated in Figure \ref{result:many}-(a), Figure \ref{result:many}-(b), respectively.

\begin{table}[htb!]
\centering
\caption{Configuration of the training and test for the car model recognition.}
\label{car_model_set}
\resizebox{0.49\textwidth}{!}{
\begin{tabular}{|c|c|c|c|c|}

\hline
\textbf{\# (Number) of} & \multicolumn{3}{|c|}{\textbf{Training Set Options}} & \textbf{Test} \\\cline{2-4}
\textbf{images} & \textbf{Model-10\%} & \textbf{Model-50\%} & \textbf{Model-100\%} & \textbf{Set} \\
\hline

\textbf{per class} & 100 & 500 & 1,000 & 500 \\
\textbf{Total \# (2-class)} & 200 & 1,000 & 2,000 & 1,000 \\
\textbf{Total \# (10-class)} & 1,000 & 5,000 & 10,000 & 5,000 \\

\hline

\end{tabular}
}
\end{table}

\subsubsection{Experiment Results}
Following the experiment results shown in Figure \ref{result:many}-(a), Figure \ref{result:many}-(b), we discovered several takeaways regarding the effectiveness of pre-training. First, as a response to the \textbf{RQ1}, every pre-training method (\textit{TL-ImageNet}, \textit{TL-Car}s, \textit{SSL-Rotation}) accomplished superior accuracy than random initialization (\textit{Random}) regardless of training set options. We analyzed this escalation of target task accuracy concretely derives from the acquired representation power during the pre-training, and examined the pre-training methods are valid in the car model recognition task. Furthermore, as a response to the \textbf{RQ2}, we discovered the magnitude of target task accuracy escalation differs along with the size of the target training set. Given the small size of the target training set (\textit{Model-10\%}), \textit{SSL-Rotation} was the most effective one among pre-training methods. Along with the increased size of the training set (from \textit{Model-10\%} option to the \textit{Model-50\%}, \textit{Model-100\%}), \textit{SSL-Rotation} consistently accomplishes the best target task accuracy, but the other pre-training methods also achieve similar performance to it. 

\subsection{Many-shot Car Defect Recognition}
\subsubsection{Setup}
In this section, we examined whether the takeaways discovered in section \ref{many:model} are also valid in the car defect recognition task, which is a binary classification between the normal and damaged images. In response to the \textbf{RQ1}, we employed four pre-training options presented in the section\ref{options}, but additionally utilized one extra pre-training option denoted as \textit{TL-Cars-OURS}, which is the fine-tuned network on 10-class car model recognition. We analyzed the trained neural networks in the car model recognition task would yield a particular amount of knowledge to the car defect recognition; thus, the trained feature extractor $F_T$ at car model recognition can become a trained source feature extractor $F_S$ at the car defect recognition. Note that we  regard the \textit{car model dataset} as a source dataset $D_S$ and \textit{car defect recognition} as a target dataset $D_T$ in this experiment. For the \textit{TL-Cars-OURS} pre-training method, we utilized the feature extractor of the car model recognition model trained under 10-class classification setting with \textit{Model-100\%} training set option and \textit{TL-ImageNet} pre-trained weight. In response to the \textbf{RQ2}, as shown in Table \ref{car_defect_dataset}, we also configured three training set options (\textit{Defect-10\%}, \textit{Defect-50\%}, \textit{Defect-100\%}) consisting of different numbers of images at each class.

\begin{table}[htb!]
\centering
\caption{Configuration of the car defect recognition.}
\label{car_defect_dataset}
\resizebox{0.45\textwidth}{!}{
\begin{tabular}{|c|c|c|c|c|}

\hline
\textbf{\# (Number) of} & \multicolumn{3}{|c|}{\textbf{Training Set Options}} & \textbf{Defect} \\\cline{2-4}
\textbf{images at} & \textbf{Defect-10\%} & \textbf{Defect-50\%} & \textbf{Defect-100\%} & \textbf{-Test} \\
\hline

\textbf{Normal Class} & 296 & 1,474 & 2,946 & 685 \\
\textbf{Damaged Class} & 122 & 611 & 1,222 & 700 \\
\hline
\textbf{Total} & 418 & 2,085 & 4,168 & 1,385 \\

\hline

\end{tabular}
}
\end{table}


As an implementation detail, we utilized the same configurations presented in section \ref{many:model} except for the neural networks architecture. In car defect recognition, we employed Progressive Multi-Granularity nerual networks (PMG-Net) \cite{du2020fine}, which achieved a precise performance in fine-grained classification task \cite{zhao2017survey,wei2019deep}. Compared to the car model recognition task, we analyzed that the car defect recognition task requires the model to capture more finer-grained discriminative characteristics between normal and damaged car images. 
Therefore, we employed PMG-Net as a neural networks architecture in this car defect recognition as it is originally designed to capture fine-grained discriminative characteristics between classes similar to each other. Note that we tried to solve the car defect recognition task with ResNet-50 architecture, but it failed to converge during the training stage under various hyperparameters. Following the experiment setups, the results are shown in Figure \ref{result:many}-(c).

\subsubsection{Experiment Result}
Referring to the experiment result shown in Figure \ref{result:many}-(c), we figured out the proposed takeaways at car model recognition are also valid in car defect recognition, which requires the model to understand fine-grained discriminative characteristics among each class. As a response to the \textbf{RQ1}, every pre-training method accomplished superior target accuracy rather than random initialization. Furthermore, as a response to the \textbf{RQ2}, the self-supervised learning (\textit{SSL-Rotation}) method accomplished precise target accuracy regardless of the training set options.

While every pre-training method accomplished similar target task accuracy in car model recognition task, only SSL-Rotation and TL-ImageNet achieved similar, superior performance under the large size of the target training set (\textbf{Defect-100\%}). We analyzed these results derived from the biasedness of the knowledge learned during the pre-training. Both TL-Cars and TL-Cars-OURS methods convey the knowledge biased to the classes in the source dataset $D_S$. TF-Cars and TF-Cars-OURS pre-train the representation power to extract meaningful features biased to car models biased to 196 car models and 10 car models, respectively. As discriminative characteristics among car classes are less fine-grained than car defect recognition, we analyzed the representation power biased to the coarse-grained characteristics (learned at TL-Cars and TL-Cars-OURS) could not provide a significant knowledge for the car defect recognition, which is a finer-grained recognition task. On the other hand, we interpret TL-ImageNet and SSL-Rotation methods yield comparatively less biased knowledge to the target task. We analyzed TL-ImageNet is less biased to the classes in the source dataset as there exist a thousand classes, and SSL-Rotation is also less biased due to its inherent characteristics \cite{jing2020self,jaiswal2021survey}. Note that self-supervised learning is inherently designed to evade the convey of biased representation power. 
We analyzed this less biased representation power conveys better knowledge to the target feature extractor on fine-grained image recognition tasks; thus, TL-ImageNet and SSL-Rotation consistently could increase the target task accuracy rather than other pre-training methods.


    
    
    

\section{Effectiveness of Pre-Training under Few-shot setting} \label{sec5}

\subsection{Problem Definition}
In this section, we validated how the pre-training methods elevate the target task performance under the few-shot learning paradigm. The few-shot learning paradigm aims to recognize classes with a few labeled samples at the target dataset. 
With the conventional case, training the target network on this dataset presumably causes overfitting as the few samples cannot provide sufficient knowledge to the model; thus, the model cannot achieve satisfactory performance. 
Instead, the few-shot learning employs a meta-learning stage with an external auxiliary task to let the model understand how to extract features from the image. During the training phase, the few-shot learning paradigm first trains the neural networks with a labeled, easily acquirable auxiliary dataset consisting of many samples. Then, the trained neural networks adapt to the novel classes (the target classes) during the test phase, where a few samples exist at each novel class \cite{bateni2020improved,koch2015siamese}. As the few-shot learning paradigm does not necessitate many labeled samples in the target dataset, practitioners utilize this paradigm when they cannot acquire the labeled target dataset. Under the scope of our study at the few-shot learning paradigm, we presume the auxiliary dataset as \textit{car model dataset} and the target dataset as \textit{car defect dataset}. While \textit{car model dataset} can be easily acquired from the operations or web search, we presumed collecting damaged car images reflecting real-world damages (i.e., various types of scratch, dent, separated car parts) is more challenging to the practitioners. To validate the effectiveness of pre-training in the few-shot learning paradigm, we also established a research question as below.

\begin{itemize}
    \item \textbf{RQ3:} Does the pre-training contribute to the escalation of target task performance?
\end{itemize}

\begin{table}[htb!]
\centering
\caption{Configuration of datasets in the few-shot setting}
\label{few_shot_dataset}
\resizebox{0.40\textwidth}{!}{
\begin{tabular}{|c|c|c|c|}

\hline
\textbf{Descriptions of} & \multicolumn{3}{|c|}{\textbf{Dataset for Few-Shot Classification}} \\\cline{2-4}
\textbf{the dataset} & \textbf{Auxiliary Set} & \textbf{Support Set} & \textbf{Query Set}  \\
\hline

\textbf{Source} & Model-100\% & Defect-100\% & Defect-Test \\
\textbf{Label Space} &6-class & 2-class & 2-class \\
\textbf{Total \# of images} & 6,000 & 4,168 & 1,385 \\
\hline

\end{tabular}
}
\end{table}

\subsection{Few-Shot Car Defect Recognition} \label{sec:few}
\subsubsection{Setup}
The few-shot learning paradigm bears two stages: 1) training stage with the auxiliary set (which is \textit{car model dataset}), 2) test stage with the support set and the test set (where both support set and test set are derived from the target task, which is \textit{car defect dataset}). The auxiliary set is a set of data utilized to let the neural networks understand how to extract the knowledge at given images. The support set and query set consists of the unseen classes during the training stage; thus, their label space differs from the auxiliary set. The support set includes a few labeled samples from the target task. It is utilized as a guideline to the trained model to infer discriminative characteristics of the target task, which consists of novel classes in the model's perspective. Lastly, the query set bears a particular amount of samples from the target task, which is utilized for the model validation during the test stage. If the support set includes $K$ samples for each of $N$ classes, we conventionally denote this few-shot recognition as $N$-way $K$-shot setting. In this study, we established the few-shot recognition as a 2-way 5-shot setting (2-way derives from two classes exist in the \textit{car defect dataset}). Considering the aforementioned descriptions regarding the few-shot learning paradigm, we configured auxiliary dataset, support set, and a query set as shown in Table \ref{few_shot_dataset}. Note that we utilized 6 randomly selected classes from \textit{car model dataset} to reduce the burden of computation.

Among numerous few-shot classification approaches\cite{snell2017prototypical,finn2017model}, we employed Relation Network proposed in \cite{sung2018learning} due to its precise performance in the public benchmark dataset. Given a query image, Relation Network identifies the most similar class in the support set by selecting the smallest distance between support samples and a query sample. 
To examine the effectiveness of pre-training at a few-shot classification setting, we first initialized the feature extractor of the target network ($N_T$) with five options: Random, TL-ImageNet, TL-Cars, TL-Cars-OURS, and SSL-Rotation. Note that source labels in the TL-Cars-OURS option do not duplicate the labels that exist in the auxiliary set. We then let the target network be trained on the auxiliary set and examined the few-shot classification performance with the support and query sets. As the target task performance can vary along with the selected samples in the support set, we applied the episode learning method \cite{snell2017prototypical}, which averages the test accuracy at $E$ iterations of the test stage. In a single iteration, episode learning randomly selects five support samples at each class of support set and calculates the classification accuracy; thus, the averaged accuracy can measure an overall accuracy regardless of the selected support samples. In this study, we recorded the target task accuracy scores within 90\% of the confidence interval from 100 iterations. The experiment result is shown in Table \ref{table:few-shot}.

\begin{table}[htb!]
\centering
\caption{2-way 5-shot car damage recognition performance 
}
\resizebox{0.40\textwidth}{!}{
\begin{tabular}{|c|c|c|c|}

\hline
\textbf{Pre-Training} & \multicolumn{3}{|c|}{\textbf{Accuracy(\%)}} \\\cline{2-4}
\textbf{Methods} & \textbf{Interval} & \textbf{Lower Bound} & \textbf{Upper Bound}  \\
\hline

\textbf{Random} &53.38$\pm$5.2& 48.63 & 59.03 \\
\textbf{TL-ImageNet} &75.59$\pm$ 2.2& 73.39 & 77.79 \\
\textbf{TL-Cars} &75.16$\pm$2.3& 72.86 & 77.46 \\
\textbf{SSL-Rotation} &\textbf{75.91}$\pm$\textbf{2.4}& \textbf{73.51} & \textbf{78.31} \\
\hline
\end{tabular}
} 

\label{table:few-shot}
\end{table}

\subsubsection{Experiment Result}
Following the experiment result elaborated in Table \ref{table:few-shot}, we figured out several takeaways similar to the lessons learned in experiments under the many-shot setting. In response to the \textbf{RQ3}, we discovered that the pre-training method contributed to the escalated target accuracy rather than the random initialization. We interpret pre-training methods conveyed a particular amount of representation power to the target neural network to extract meaningful features in a given image. Moreover, while the self-supervised learning method achieved the best target task performance, there exists a minimal difference among various pre-training methods. Therefore, we expect candidate practitioners can utilize any pre-training method in the real-world problem to escalate the few-shot classification performance.


    

\begin{figure*}[ht]
\centering
\begin{tabular} {c c} \\
\includegraphics[width=0.40\textwidth, trim={0cm 0cm 0cm 0cm}, clip]{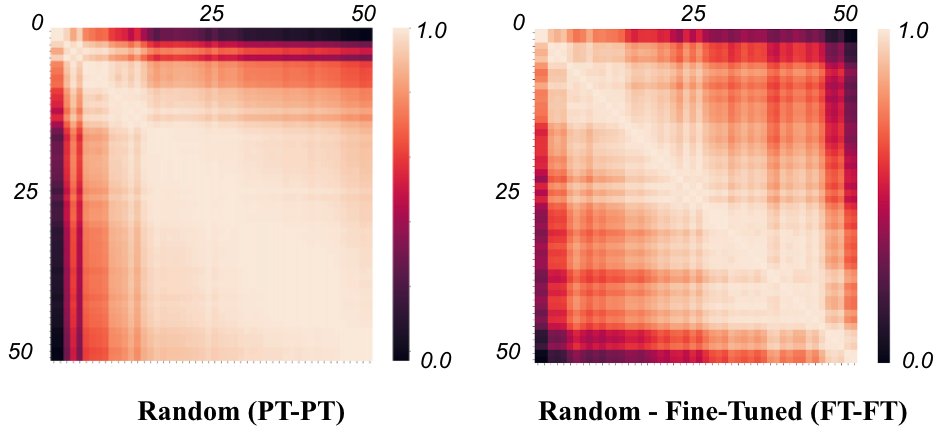} & \includegraphics[width=0.40\textwidth, trim={0cm 0cm 0cm 0cm}, clip]{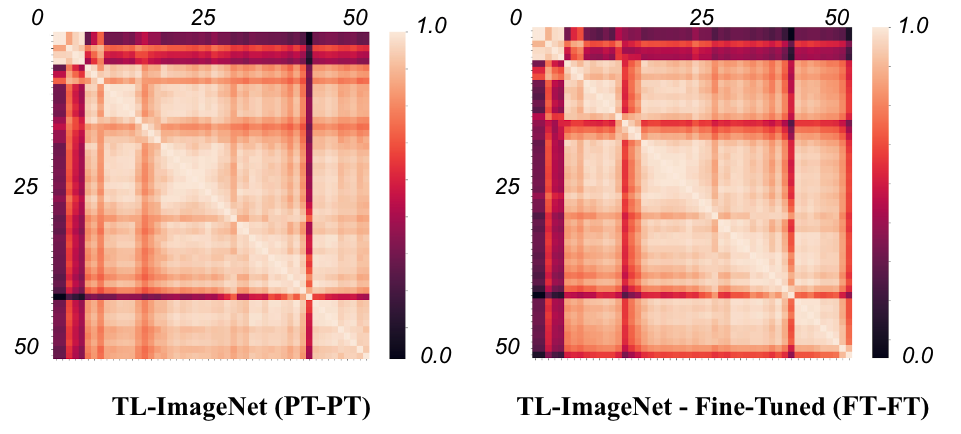}\\
(a) Random & (b) TL-ImageNet \\

\\ 

\includegraphics[width=0.40\textwidth, trim={0cm 0cm 0cm 0cm}, clip]{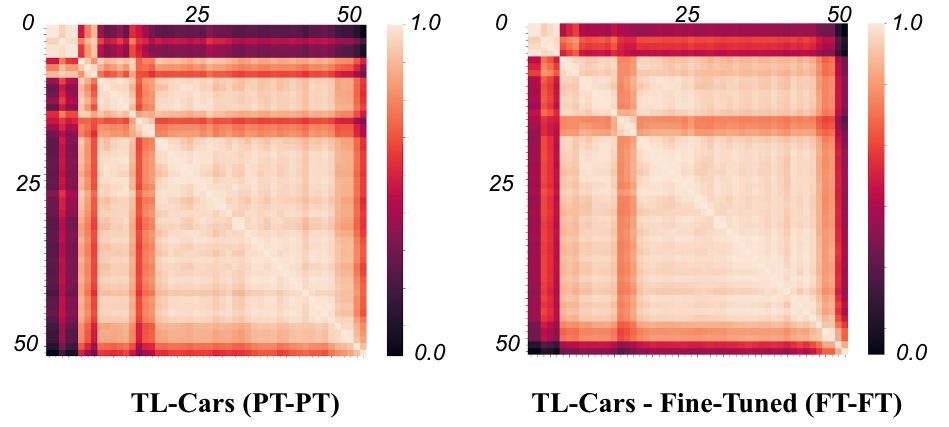} & \includegraphics[width=0.40\textwidth, trim={0cm 0cm 0cm 0cm}, clip]{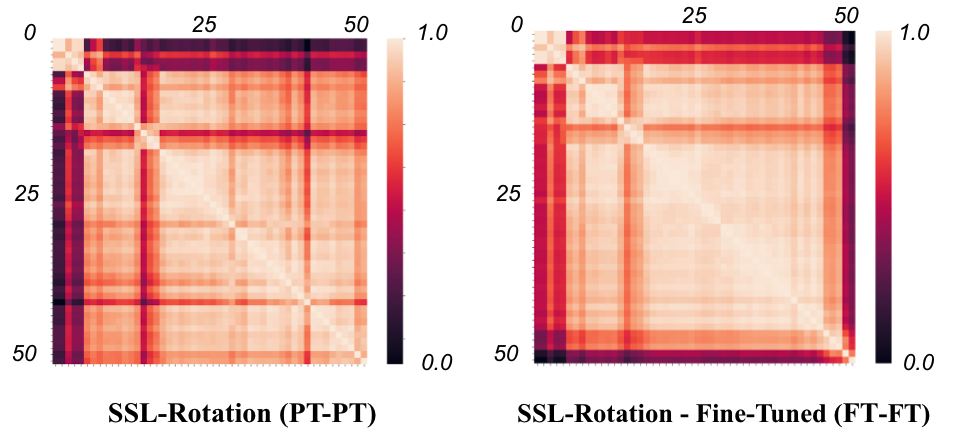}\\
(c) TL-Cars & (d) SSL-Rotation \\
\end{tabular}
\caption{Layer-wise similarity matrices in a single neural network. 
}
\label{fig:cka_viz}
\end{figure*}

\section{How does the Pre-Training escalate target task performance?}
Throughout experimental analyses in section \ref{sec:many} and \ref{sec:few}, we discovered pre-training methods concretely contribute to the escalated target task performance in every image recognition task and problem settings. Based on the aforementioned discoveries, we aim to scrutinize an underlying reason behind the effectiveness of pre-training by analyzing learned knowledge at the neural networks. To analyze the representations in neural networks consisting of different weights, we employed Central Kernel Alignment (CKA) \cite{kornblith2019similarity} as an index of measuring similarity between two representations from different neural networks. Suppose we provide image samples into two neural networks ($N_1$, $N_2$) trained from different pre-trained weights. Then, we can extract a pair of representation vectors from any layers of $N_1$ and $N_2$ denoted as ($R_1$, $R_2$). Given these representation vectors $R_1$ and $R_2$, CKA effectively measures the similarity between layers in the same neural networks with different weights and across entirely different architectures. The CKA yields a similarity metric lying between 0 to 1, where 0 implies less similarity and 1 illustrates high similarity. Due to its convenience and effectiveness in measuring similarity between two representations, we utilized it. Please refer to the original publication \cite{kornblith2019similarity} for a detailed elaboration on the CKA.

To scrutinize the reason behind the superior contribution of various pre-training methods, we established the following research questions.

\begin{itemize}
    \item \textbf{RQ 4}: What does the pre-trained weight convey to the target task? Does the pre-trained weight help the neural networks understand data a priori to the target task training?
    
    \item \textbf{RQ 5}: Which layers of the network are primarily affected by the pre-training and the fine-tuning?
\end{itemize}

\begin{table}[htb!]
\centering
\caption{Representation similarity in lower-level blocks}
\resizebox{0.35\textwidth}{!}{
\begin{tabular}{|c|c|c|c|}

\hline
\textbf{Pre-Training} & \multicolumn{3}{|c|}{\textbf{Representation Similarity}} \\\cline{2-4}
\textbf{Methods} & \textbf{(PT, PT)} & \textbf{(PT, FT)} & \textbf{Difference}  \\
\hline

\textbf{TL-ImageNet} & 0.9192 & 0.9212 & 0.0020 \\
\textbf{TL-Cars} & 0.9447 & 0.9336 & 0.0111 \\
\textbf{SSL-Rotation} & 0.9224 & 0.9240 & 0.0016 \\
\hline

\end{tabular}
}
\label{table:cka_low}
\end{table}

\begin{table}[htb!]
\centering
\caption{Representation similarity in higer-level blocks}
\resizebox{0.35\textwidth}{!}{
\begin{tabular}{|c|c|c|c|}

\hline
\textbf{Pre-Training} & \multicolumn{3}{|c|}{\textbf{Representation Similarity}} \\\cline{2-4}
\textbf{Methods} & \textbf{(PT, PT)} & \textbf{(PT, FT)} & \textbf{Difference}  \\
\hline

\textbf{TL-ImageNet} & 0.9819 & 0.8526 & 0.1293 \\
\textbf{TL-Cars} & 0.9579 & 0.8548 & 0.1031 \\
\textbf{SSL-Rotation} & 0.9757 & 0.7942 & 0.1815 \\
\hline

\end{tabular}
}
\label{table:cka_high}
\end{table}

\subsection{Analyis on the Overall Knowledge}

In response to the \textbf{RQ 4}, we visualized layer-wise similarity among every layer of the neural networks. We would like to highlight that this section employs four pre-trained weights proposed in \textcolor{black}{section \ref{sec5}}, and target network (ResNet-50) trained on the 2-class car model dataset (Model-100\%). Given a set of samples at \textit{Model-Test}, we provide these samples into the neural network and acquire the representations at $i$th block and $j$th block, denoted as $R_i$ and $R_j$, respectively. We measured the representation similarity between $R_i$ and $R_j$, and created a similarity matrix consists of $(i,j)$.
Just as \textcolor{black}{\cite{kornblith2019similarity}} utilized this similarity matrix as a proxy to overall knowledge learned by the neural network, we also employed it in the analysis. To scrutinize the impact of pre-training on the target task, we compared the similarity matrix between the pre-trained feature extractor and the fine-tuned feature extractor. Suppose the similarity matrix of the pre-trained feature extractor looks similar to the fine-tuned feature extractor. In that case, we can infer that the pre-training conveys a particular amount of knowledge to the target neural network before the target training stage. Following these setups, we visualized a pair of similarity matrices with four pre-training options in Figure \ref{fig:cka_viz}. 

Referring to Figure \ref{fig:cka_viz}, we figured out several interesting takeaways regarding the effectiveness of pre-training. First, similarity matrices of TL-ImageNet, TL-Cars, SSL-Rotation share similar patterns while each pre-trained weight was established from a different source dataset. (Refer the similarity matrices denoted as PT-PT). Unlike the similarity matrix at the randomly initialized network, three pre-trained weights (PT-PT) shared a similar shape of a square-like structure. Accordingly, we analyzed the existence of common knowledge among each pre-trained weight, and this knowledge provides a precedent knowledge to the target neural network. Second, we figured out the fine-tuned network also maintains similar knowledge to the pre-trained weights. Comparing each pair of (PT-PT, PT-FT) in Figure \ref{fig:cka_viz}, a pair consisting of pre-trained weights seems to be similar while the similarity matrix at random initialization (PT-PT) changed in a particular manner at PT-FT. We analyzed the target task training provides additional knowledge to the neural networks based on the common knowledge from the pre-training. As the amount of change from PT-PT to PT-FT is not magnificent, we also expect target task training to yield a smaller amount of knowledge to the neural networks than pre-training. Throughout the analysis on the overall knowledge in the neural networks, we inferred that various pre-training methods establish common knowledge regarding image understanding, and it becomes a solid basis for the target neural network to accomplish precise target task performance.

\subsection{Analysis on the Layer-level Knowledge=}

While we figured out the pre-training method concretely yields a common knowledge to the target neural network, we scrutinized which block (layers at conventional CNN architectures) of the neural network primarily illustrates a common knowledge conveyed from the pre-trained weight. Not only the conveyed common knowledge, but we also scrutinized which block of the neural network includes additional knowledge acquired during the fine-tuning into the target task. As a response to the \textbf{RQ5}, we compared the similarity between the pre-trained weight (PT) and the fine-tuned weight (PT-FT) at two blocks: low-level block and high-level block. A low-level block is the first block at ResNet-50 architecture, which first receives the image as an input and extracts low-level characteristics of the image. A high-level block is the last block at ResNet-50, where yields a condensed representation to the classification layers. We scrutinized how does the knowledge changes during the fine-tuning stage by comparing the representation similarity. The representation similarity between PT and PT-FT at low-level block and high-level block are illustrated in Table \ref{table:cka_low}, Table \ref{table:cka_high}, respectively. 

We resulted in the common knowledge conveyed by the pre-trained weight primarily exist in the low-level block, while the knowledge gained from the fine-tuning stage lies in the high-level block. As shown in Table \ref{table:cka_low}, representations after the fine-tuning do not change in a particular manner. Therefore, we expect a common knowledge generated by the pre-training presumably exists in the neural network's low-level blocks, which implies the pre-training conveys the target network how to extract low-level features. Referring to Table \ref{table:cka_high}, we discovered representation similarity at the high-level block decreases in a particular manner after the fine-tuning; thus, we infer the fine-tuning provides knowledge regarding the semantics of the target dataset.






\section{Conclusion}

Our study performed in-depth analyses of which learning paradigm fits with real-world car images and how the representation landscape looks in the trained neural networks. 
First, we highly recommend that practitioners utilize the pre-training method as they accomplish improved target task performances rather than random initialization in every image recognition task. Second, the practitioners shall be cautious about maintaining unbiasedness in the pre-trained weight under the fine-grained target task. Lastly, we emphasize that the conveyed representation power from the pre-training primarily exists in the low-level layers of the neural networks, while the fine-tuning provides knowledge in the high-level layers. Still, we acknowledge there exist several improvement avenues for our study. Further studies shall compare various self-supervised learning methods or compare the effectiveness of transfer learning from non-natural images. 

\bibliographystyle{IEEEtranS.bst}
\bibliography{main}

\begin{thebibliography}{10}
\providecommand{\url}[1]{#1}
\csname url@samestyle\endcsname
\providecommand{\newblock}{\relax}
\providecommand{\bibinfo}[2]{#2}
\providecommand{\BIBentrySTDinterwordspacing}{\spaceskip=0pt\relax}
\providecommand{\BIBentryALTinterwordstretchfactor}{4}
\providecommand{\BIBentryALTinterwordspacing}{\spaceskip=\fontdimen2\font plus
\BIBentryALTinterwordstretchfactor\fontdimen3\font minus
  \fontdimen4\font\relax}
\providecommand{\BIBforeignlanguage}[2]{{%
\expandafter\ifx\csname l@#1\endcsname\relax
\typeout{** WARNING: IEEEtranS.bst: No hyphenation pattern has been}%
\typeout{** loaded for the language `#1'. Using the pattern for}%
\typeout{** the default language instead.}%
\else
\language=\csname l@#1\endcsname
\fi
#2}}
\providecommand{\BIBdecl}{\relax}
\BIBdecl

\bibitem{azimjonov2021real}
J.~Azimjonov and A.~{\"O}zmen, ``A real-time vehicle detection and a novel
  vehicle tracking systems for estimating and monitoring traffic flow on
  highways,'' \emph{Advanced Engineering Informatics}, vol.~50, p. 101393,
  2021.

\bibitem{azimjonov2022vision}
------, ``Vision-based vehicle tracking on highway traffic using bounding-box
  features to extract statistical information,'' \emph{Computers \& Electrical
  Engineering}, vol.~97, p. 107560, 2022.

\bibitem{bateni2020improved}
P.~Bateni, R.~Goyal, V.~Masrani, F.~Wood, and L.~Sigal, ``Improved few-shot
  visual classification,'' in \emph{Proceedings of the IEEE/CVF Conference on
  Computer Vision and Pattern Recognition}, 2020, pp. 14\,493--14\,502.

\bibitem{caron2020unsupervised}
M.~Caron, I.~Misra, J.~Mairal, P.~Goyal, P.~Bojanowski, and A.~Joulin,
  ``Unsupervised learning of visual features by contrasting cluster
  assignments,'' \emph{arXiv preprint arXiv:2006.09882}, 2020.

\bibitem{chen2020simple}
T.~Chen, S.~Kornblith, M.~Norouzi, and G.~Hinton, ``A simple framework for
  contrastive learning of visual representations,'' in \emph{International
  conference on machine learning}.\hskip 1em plus 0.5em minus 0.4em\relax PMLR,
  2020, pp. 1597--1607.

\bibitem{datondji2016survey}
S.~R.~E. Datondji, Y.~Dupuis, P.~Subirats, and P.~Vasseur, ``A survey of
  vision-based traffic monitoring of road intersections,'' \emph{IEEE
  transactions on intelligent transportation systems}, vol.~17, no.~10, pp.
  2681--2698, 2016.

\bibitem{deng2009imagenet}
J.~Deng, W.~Dong, R.~Socher, L.-J. Li, K.~Li, and L.~Fei-Fei, ``Imagenet: A
  large-scale hierarchical image database,'' in \emph{2009 IEEE conference on
  computer vision and pattern recognition}.\hskip 1em plus 0.5em minus
  0.4em\relax Ieee, 2009, pp. 248--255.

\bibitem{du2020fine}
R.~Du, D.~Chang, A.~K. Bhunia, J.~Xie, Z.~Ma, Y.-Z. Song, and J.~Guo,
  ``Fine-grained visual classification via progressive multi-granularity
  training of jigsaw patches,'' in \emph{European Conference on Computer
  Vision}.\hskip 1em plus 0.5em minus 0.4em\relax Springer, 2020, pp. 153--168.

\bibitem{fang2016fine}
J.~Fang, Y.~Zhou, Y.~Yu, and S.~Du, ``Fine-grained vehicle model recognition
  using a coarse-to-fine convolutional neural network architecture,''
  \emph{IEEE Transactions on Intelligent Transportation Systems}, vol.~18,
  no.~7, pp. 1782--1792, 2016.

\bibitem{finn2017model}
C.~Finn, P.~Abbeel, and S.~Levine, ``Model-agnostic meta-learning for fast
  adaptation of deep networks,'' in \emph{International Conference on Machine
  Learning}.\hskip 1em plus 0.5em minus 0.4em\relax PMLR, 2017, pp. 1126--1135.

\bibitem{gidaris2018unsupervised}
S.~Gidaris, P.~Singh, and N.~Komodakis, ``Unsupervised representation learning
  by predicting image rotations,'' \emph{arXiv preprint arXiv:1803.07728},
  2018.

\bibitem{grill2020bootstrap}
J.-B. Grill, F.~Strub, F.~Altch{\'e}, C.~Tallec, P.~Richemond, E.~Buchatskaya,
  C.~Doersch, B.~Avila~Pires, Z.~Guo, M.~Gheshlaghi~Azar \emph{et~al.},
  ``Bootstrap your own latent-a new approach to self-supervised learning,''
  \emph{Advances in neural information processing systems}, vol.~33, pp.
  21\,271--21\,284, 2020.

\bibitem{he2016deep}
K.~He, X.~Zhang, S.~Ren, and J.~Sun, ``Deep residual learning for image
  recognition,'' in \emph{Proceedings of the IEEE conference on computer vision
  and pattern recognition}, 2016, pp. 770--778.

\bibitem{jaiswal2021survey}
A.~Jaiswal, A.~R. Babu, M.~Z. Zadeh, D.~Banerjee, and F.~Makedon, ``A survey on
  contrastive self-supervised learning,'' \emph{Technologies}, vol.~9, no.~1,
  p.~2, 2021.

\bibitem{jing2020self}
L.~Jing and Y.~Tian, ``Self-supervised visual feature learning with deep neural
  networks: A survey,'' \emph{IEEE Transactions on Pattern Analysis and Machine
  Intelligence}, 2020.

\bibitem{khan2020survey}
A.~Khan, A.~Sohail, U.~Zahoora, and A.~S. Qureshi, ``A survey of the recent
  architectures of deep convolutional neural networks,'' \emph{Artificial
  Intelligence Review}, vol.~53, no.~8, pp. 5455--5516, 2020.

\bibitem{kingma2014adam}
D.~P. Kingma and J.~Ba, ``Adam: A method for stochastic optimization,''
  \emph{arXiv preprint arXiv:1412.6980}, 2014.

\bibitem{koch2015siamese}
G.~Koch, R.~Zemel, R.~Salakhutdinov \emph{et~al.}, ``Siamese neural networks
  for one-shot image recognition,'' in \emph{ICML deep learning workshop},
  vol.~2.\hskip 1em plus 0.5em minus 0.4em\relax Lille, 2015.

\bibitem{kornblith2019similarity}
S.~Kornblith, M.~Norouzi, H.~Lee, and G.~Hinton, ``Similarity of neural network
  representations revisited,'' in \emph{International Conference on Machine
  Learning}.\hskip 1em plus 0.5em minus 0.4em\relax PMLR, 2019, pp. 3519--3529.

\bibitem{KrauseStarkDengFei-Fei_3DRR2013}
J.~Krause, M.~Stark, J.~Deng, and L.~Fei-Fei, ``3d object representations for
  fine-grained categorization,'' in \emph{4th International IEEE Workshop on 3D
  Representation and Recognition (3dRR-13)}, Sydney, Australia, 2013.

\bibitem{krogh1992simple}
A.~Krogh and J.~A. Hertz, ``A simple weight decay can improve generalization,''
  in \emph{Advances in neural information processing systems}, 1992, pp.
  950--957.

\bibitem{noroozi2016unsupervised}
M.~Noroozi and P.~Favaro, ``Unsupervised learning of visual representations by
  solving jigsaw puzzles,'' in \emph{European conference on computer
  vision}.\hskip 1em plus 0.5em minus 0.4em\relax Springer, 2016, pp. 69--84.

\bibitem{pan2009survey}
S.~J. Pan and Q.~Yang, ``A survey on transfer learning,'' \emph{IEEE
  Transactions on knowledge and data engineering}, vol.~22, no.~10, pp.
  1345--1359, 2009.

\bibitem{park2022don}
K.~H. Park and H.~Chung, ``Don’t wait until the accident happens: Few-shot
  classification framework for car accident inspection in a real world,'' in
  \emph{International Conference on Image Analysis and Processing}.\hskip 1em
  plus 0.5em minus 0.4em\relax Springer, 2022, pp. 560--571.

\bibitem{park2021visual}
K.~H. Park, Y.~Kwon, Y.~Song, and S.~Byeon, ``Visual representation learning
  for automating car part recognition in a large-scale car sharing platform,''
  in \emph{2021 IEEE 17th International Conference on Automation Science and
  Engineering (CASE)}.\hskip 1em plus 0.5em minus 0.4em\relax IEEE, 2021, pp.
  1104--1110.

\bibitem{patil2017deep}
K.~Patil, M.~Kulkarni, A.~Sriraman, and S.~Karande, ``Deep learning based car
  damage classification,'' in \emph{2017 16th IEEE international conference on
  machine learning and applications (ICMLA)}.\hskip 1em plus 0.5em minus
  0.4em\relax IEEE, 2017, pp. 50--54.

\bibitem{singh2019automating}
R.~Singh, M.~P. Ayyar, T.~V.~S. Pavan, S.~Gosain, and R.~R. Shah, ``Automating
  car insurance claims using deep learning techniques,'' in \emph{2019 IEEE
  Fifth International Conference on Multimedia Big Data (BigMM)}.\hskip 1em
  plus 0.5em minus 0.4em\relax IEEE, 2019, pp. 199--207.

\bibitem{snell2017prototypical}
J.~Snell, K.~Swersky, and R.~S. Zemel, ``Prototypical networks for few-shot
  learning,'' \emph{arXiv preprint arXiv:1703.05175}, 2017.

\bibitem{sung2018learning}
F.~Sung, Y.~Yang, L.~Zhang, T.~Xiang, P.~H. Torr, and T.~M. Hospedales,
  ``Learning to compare: Relation network for few-shot learning,'' in
  \emph{Proceedings of the IEEE conference on computer vision and pattern
  recognition}, 2018, pp. 1199--1208.

\bibitem{wei2019deep}
X.-S. Wei, J.~Wu, and Q.~Cui, ``Deep learning for fine-grained image analysis:
  A survey,'' \emph{arXiv preprint arXiv:1907.03069}, 2019.

\bibitem{zhao2017survey}
B.~Zhao, J.~Feng, X.~Wu, and S.~Yan, ``A survey on deep learning-based
  fine-grained object classification and semantic segmentation,''
  \emph{International Journal of Automation and Computing}, vol.~14, no.~2, pp.
  119--135, 2017.

\end{thebibliography}

\end{document}